\pdfoutput=1

\documentclass[11pt]{article}

\usepackage[]{authblk}
\usepackage{mdframed}
\usepackage{setspace}
\usepackage{multirow}
\usepackage{amsmath} 
\usepackage{multirow}
\usepackage{booktabs}
\usepackage[utf8x]{inputenc}
\usepackage{cjhebrew}
\usepackage{hhline}
\usepackage{arydshln}

\makeatletter
\renewcommand\AB@affilsepx{, \protect\Affilfont}
\makeatother

\usepackage[]{ACL2023}
\let\oldfootnotetext\footnotetext
\renewcommand{\footnotetext}[1]{%
  \begingroup%
  \renewcommand{\thefootnote}{\ensuremath{*}}%
  \oldfootnotetext{#1}%
  \endgroup%
}

\usepackage{times}
\usepackage{latexsym}
\usepackage{graphicx}
\usepackage{tabu}
\usepackage{multirow}
\usepackage{makecell} 

\usepackage[T1]{fontenc}

\usepackage{microtype}

\usepackage{inconsolata}
\usepackage{csquotes}

\usepackage{todonotes}

\newcommand{\pz}{\hphantom{0}}

\title{HeSum: a Novel Dataset for Abstractive Text Summarization in Hebrew}

\author[,a]{Tzuf Paz-Argaman*}
\author[,a]{Itai Mondshine*}
\author[a]{Asaf Achi Mordechai}
\author[a]{Reut Tsarfaty}

\affil[a]{Bar-Ilan University, Israel}

\affil[ ]{\authorcr \tt \{tzuf.paz-argaman,   mondshi1, asaf.achimordechai, reut.tsarfaty\}@biu.ac.il}

\begin{document}
\maketitle
\begin{abstract}

While large language models (LLMs) excel in various natural language tasks in English, their performance in lower-resourced languages like Hebrew, especially for generative tasks such as abstractive summarization, remains unclear. The high morphological richness in Hebrew adds further challenges due to the ambiguity in sentence comprehension and the complexities in  meaning construction.
In this paper, we address this  resource and evaluation gap by introducing HeSum, a novel benchmark  specifically designed for  abstractive text summarization in Modern Hebrew. HeSum consists of 10,000 article-summary pairs sourced from Hebrew news websites written by professionals. Linguistic analysis confirms HeSum's high abstractness and  unique morphological challenges. We show that HeSum presents distinct difficulties  for contemporary  state-of-the-art LLMs, establishing it as a valuable testbed for  generative language technology in Hebrew, and MRLs generative challenges in general.\footnotetext{Equal contribution.}\footnote{The dataset, code, and fine-tuned models are publicly available at
 \url{https://github.com/OnlpLab/HeSum}
 }

\end{abstract}

\section{Introduction}

Recent advances with large language models (LLMs,  \citealp{brown2020language, 
JMLR:v24:22-1144})  demonstrate impressive capabilities, encompassing diverse tasks such as natural language (NL) understanding and reasoning, including {\em classification} tasks such as commonsense reasoning \cite{bisk2020piqa} and sentiment analysis \cite{liang2022holistic}, as well as  {\em generative} tasks like summarization and dialogue systems \cite{thoppilan2022lamda}.
However, these impressive achievements are primarily demonstrated for the English language. 
Our understanding of how these models perform on low-resource languages is limited, as current evaluations are primarily focused on languages with abundant data \citep{ahuja2023mega, lai2023chatgpt}.

 This concern is particularly relevant for {\em morphologically rich languages} (MRLs) such as Hebrew, which is known for  their  word complexity and ambiguity, leading to processing difficulty \cite{tsarfaty2019s, tsarfaty2020spmrl}. 
Despite advances in natural language processing for Hebrew, which so far covered tasks as  
reading comprehension (\citealp {keren2021parashoot, cohen2023heq}), named entity recognition \cite{bareket2021neural},
sentiment analysis \cite{chriqui2022hebert}, and text-based geolocation  \cite{paz2023hegel}; 
a crucial gap persists in the ability to evaluate  novel, human-like generated text, as  in {\em abstractive text generation}.

Abstractive text-generation requires both
natural language understanding and reasoning over the input, and  the ability to generate grammatically, and in particular {\em morpho-syntactically},  correct text, as well as {\em semantically} and   {\em morpho-semantically} coherent, fluent text that conveys consistent meanings. Notably, text-generation models are also prone to `hallucinations' --- generating factually incorrect 
content.
These challenges are further amplified in Hebrew  due to its morphological richness which leads to a 
complex realization of sentence structure and meaning.

\begin{table*}[h]
\centering
\scalebox{0.73}{
\begin{tabular}{lcccccccccc}
\cline{1-11}  
\multirow{2}{*}{Set} & Size  & \multicolumn{2}{c}{\begin{tabular}[c]{@{}c@{}}Vocabulary size \\ (over Articles)\end{tabular}} &  & \multicolumn{2}{c}{\begin{tabular}[c]{@{}c@{}}Avg. Document \\ Length\end{tabular}} & \begin{tabular}[c]{@{}c@{}}Avg. Word \\ Ambiguity\end{tabular} & \begin{tabular}[c]{@{}c@{}}Avg. Morph \\ Anaphors\end{tabular} & \begin{tabular}[c]{@{}c@{}}Avg.\\ Construct-state\end{tabular} & \begin{tabular}[c]{@{}c@{}}BertScore \\ Semantic Similarity\end{tabular} \\ \cline{3-4} \cline{6-7}
                     &       & Lemmas                                        & Tokens                                         &  & Article                                  & Summary                                  & Article                                                        & Article                                                        & Summary                                                        & Article-Summary                                                          \\
Train                & 8,000 & 47,903                                        & 269,168                                        &  & 1,427.4                                  & 33.2                                     & 50                                                             & 98.8                                                           & 2.4                                                            & 76                                                                       \\
Validation           & 1,000 & 23,134                                        & 104,383                                        &  & 1,410.0                                  & 33.8                                     & 90                                                             & 87.9                                                           & 2.5                                                            & 76                                                                       \\
Test                 & 1,000 & 22,543                                        & 102,387                                        &  & 1,507.6                                  & 34.7                                     & 89                                                             & 95.7                                                           & 2.6                                                            & 74   \\ \cline{1-11}                                                              
\end{tabular}
}
\caption{Linguistic Analysis of the HeSum dataset.
}
\label{tab:dataset-stats}
\end{table*}

In order to enable empirically quantified assessment of these aspects of text generation in MRLs, we present a novel benchmark dataset for \textbf{He}brew abstractive text \textbf{Sum}marization ({\bf HeSum}).  HeSum consists of 10,000 articles paired with their corresponding summaries, all of which have been sourced from three different Hebrew news websites, all written by professional journalists.  This curated collection offers several key advantages:
(i)~\emph{High Abstractness} -- extensive linguistic analysis validates HeSum's summaries as demonstrably more abstractive even when compared to  English benchmarks.
(ii)~\emph{Unique Hebrew Challenges} -- meticulous linguistic analysis  pinpoints the inherent complexities specific to Hebrew summarization, offering valuable insights into the nuanced characteristics that differentiate it from its English counterpart. And
(iii)~\emph{Thorough LLM Evaluation} -- we conducted a comprehensive empirical analysis using state-of-the-art LLMs, demonstrating that HeSum presents unique challenges even for these contemporary models. 
By combining high abstractness, nuanced morphological complexities, and a rigorous LLM evaluation, HeSum establishes itself as a valuable resource for advancing the frontiers of abstractive text summarization in MRL settings.

\section{The Challenge}
\paragraph{Linguistic Challenges in Hebrew}
Morphologically rich languages (MRLs) pose distinct challenges for generative tasks, above and beyond morphologically impoverished ones such as English.

In MRLs, each input token can be composed of multiple lexical and functional elements, each contributing to the overall  structure and semantic meanings of the generated text. For instance, the Hebrew word `\begin{cjhebrew}\cjRL{wk/smbytnw}\end{cjhebrew}' is composed of seven morphemes: `\begin{cjhebrew}w\end{cjhebrew}' (`and'), `\begin{cjhebrew}\cjRL{k/s}\end{cjhebrew}' (`when'), `\begin{cjhebrew}m|\end{cjhebrew}' (`from'), `\begin{cjhebrew}\cjRL{h}\end{cjhebrew}' (`the'), `\begin{cjhebrew}\cjRL{byt}\end{cjhebrew}' (`house'), `\begin{cjhebrew}\cjRL{/sl}\end{cjhebrew}' (`of'), and `\begin{cjhebrew}\cjRL{'n.hnw}\end{cjhebrew}' (`us').  This has ramifications for both the understanding of MRL texts,
a process that necessitates morphological segmentation, and for generating MRL texts, requiring morphological fusion.
At comprehension, Hebrew poses an additional challenge due to its inherent ambiguity, with many tokens admitting multiple valid segmentations, e.g.,  `\begin{cjhebrew}hpqh\end{cjhebrew}' 
could be interpreted as `\begin{cjhebrew}h\end{cjhebrew}'+`\begin{cjhebrew}\cjRL{qph}\end{cjhebrew}' (`the'+`coffee'); as `\begin{cjhebrew}hpqh\end{cjhebrew}' (`orbit'); or as `\begin{cjhebrew}\cjRL{hqP}\end{cjhebrew}' + `\begin{cjhebrew}\cjRL{/sl}\end{cjhebrew}'  +
`\begin{cjhebrew}\cjRL{hy'}\end{cjhebrew}'  
(`perimeter'+`of'+`her'). 
During generation, the emergence of unseen morphological compositions, where unfamiliar  morphemes combine in  familiar ways, poses an additional challenge \cite{hofmann2021superbizarre, gueta2023explicit}.
These challenges, coupled with inherent linguistic features like morphological inflections,  construct-state nouns ({\em smixut}), and flexible word order, create a multifaceted challenge for LLMs in processing and generating  Hebrew texts.

\begin{table*}[h]
\centering
\scalebox{1}{
\begin{tabular}{lcccccccc}
\toprule
Dataset & \multicolumn{4}{c}{novel n-grams} &  CMP & RED (n=1) & RED (n=2) \\
\cmidrule{2-5}
       & n = 1 & n = 2 & n = 3 & n = 4 & & & & \\
\midrule
CNN/Daily Mail & 13.20 & 52.77 & 72.20  &  81.40&  90.90 &  13.73 & 1.10 \\
XSum   & 35.76 & 83.45 & 95.50 & 98.49 & 90.40 & \pz5.83  & 0.16 \\
HeSum  & 42.00    & 73.20   & 82.00 & 85.36 &  95.48 &  \pz4.83  & 0.10 \\
\bottomrule
\end{tabular}
}
\caption{HeSum's Intrinsic Evaluation compared to English Benchmarks (CNN/Daily Mail and XSum).}
\label{tab:Abstractness}
\end{table*}

\paragraph{The HeSum Task}
We aim to unlock the comprehension-and-generation challenge in MRL settings by first tackling the abstractive text summarization task \citep{moratanch2016survey}, here focusing on Modern Hebrew.  

Given an input document in Hebrew, specifically a news article, our goal is to generate a short, clear, Hebrew summary of the key information in the article. In contrast to {\em extractive} summarization, here novel morphosyntactic structures  need to be generated to communicate the summary.

\section{Dataset, Statistics and Analysis} 

\subsection{Data Collection}
The HeSum dataset consists of article-and-summary pairs. 
The articles were collected 
from three Hebrew news websites: “Shakuf”,\footnote{\url{https://shakuf.co.il}} “HaMakom”,\footnote{\url{https://www.ha-makom.co.il}} and “The Seventh Eye”.\footnote{\url{https://www.the7eye.org.il}} These websites focus on independent journalism, providing articles on topics such as 
government accountability, corporate influence, and environmental issues. 
Each article on these websites is accompanied by an extended subheading written by a professional editor, that serves as a summary of the content.
To ensure data quality, articles that were not in Hebrew, or ones that had particularly short summaries (i.e., the extended subheading was less than 10 tokens) were excluded from the dataset. 

\subsection{Linguistic Analysis} 

We examined the linguistic, morpho-syntactic and semantic, properties of the HeSum dataset. For the extraction of syntactic and semantic features, we used DictaBert \cite{shmidman2023dictabert}. Additionally, AlephBert \cite{seker2021alephbert}, a Hebrew monolingual BERT-based encoder model \cite{devlin2018bert}, was employed to compute semantic similarity between articles and their corresponding summaries, leveraging the BertScore method \cite{zhang2019bertscore}. 
Notably, semantic similarity was performed only on article-summary pairs within the model's 512-token limit.

Table~\ref{tab:dataset-stats} highlights the Hebrew language's multifaceted complexities as reflected in this task. 
The notable disparity in the vocabulary size between token and lemma counts underscores extensive morphological richness,  necessitating models adept at handling linguistic diversity. 
The abundance of morphological anaphoric expressions   and numerous Hebrew construct-state constructions necessitate advanced models attuned to entity relations that are expressed via Hebrew's unique morphological traits. 
Lattice analysis reveals a high degree of word ambiguity (numerous lattice paths), highlighting natural language understanding challenges and the consequent difficulty of accurate tokenization for downstream processing tasks.
The substantial document length, necessitate the use of models adept at long-form text processing.
Finally, the relatively high semantic similarity score indicates effective information distillation in the summaries.

\subsection{Summarization Intrinsic Analysis}
\label{sec:data_analysis}

To assess the challenges of the HeSum summaries we used three established metrics: 
(i)~\emph{Abstractness}: 
  the percentage of  summary novel n-grams, unseen in the article \citep{narayan2018don}.
(ii)~\emph{Compression Ratio (CMP)}: 
 the word count in summary \(S\) divided by the word count in  article \(A\): \( CMP_w(S, A) = 1 - \frac{{|S|}}{{|A|}} \). 
Higher compression ratios indicate greater word-level reduction and, subsequently, potentially  a more challenging summarization task \citep{bommasani2020intrinsic}.
(iii) \emph{Redundancy (RED)}: 
measures repetitive n-grams within a summary (S) using the form:   \( RED(S) = \frac{{\sum_{i=1}^{m} (f_i - 1)}}{{\sum_{i=1}^{m} f_i}}\) where  \( m \) is the number of unique n-grams in the summary and \( f_i\geq 1 \) is the frequency count of a specific n-gram
\citep{hasan2021xl}.

\begin{table*}
\centering
\scalebox{0.9}{
\begin{tabular}{lcccccccc}
\toprule
Model & \multicolumn{3}{c}{ROUGE} & BertScore  &  \multicolumn{2}{c}{Human Evaluation}\\
\cmidrule(lr){2-4} \cmidrule(lr){6-7}
 & ROUGE-1 & ROUGE-2 & ROUGE-L   & & Coherence & Completeness\\
\midrule
GPT-4   & 13.59 & 3.70 & 10.39 & 77.3 &4.48  & 4.14 \\
GPT-3.5 & 13.69 & 3.84 & 10.55  &  77.0 &  4.38 & 3.98 \\
mLongT5 (fine-tuned)  & 17.47    & 7.56   & 14.68 & 57.6  & 3.46 & 2.10  \\

\bottomrule
\end{tabular}}
\caption{Models' performance on the HeSum test-set.}
\label{tab:models_results}
\end{table*}

Table~\ref{tab:Abstractness} presents a quantitative analysis of HeSum's summarization characteristics, underscoring its challenges. HeSum demonstrates a high degree of abstractness, with approximately half of its unique vocabulary and over 73\% of bigrams unseen in the original articles.
Furthermore, HeSum presents a significant compression challenge, as summaries average less than 5\% of the input article length. Additionally, the analysis reveals minimal redundancy within the summaries, with less than 5\% repeated n-grams.
These findings underscore HeSum's efficacy in conveying the central ideas of the articles' information in a novel, distillate, and non-redundant manner. Comparative analysis with established abstractive summarization benchmarks, CNN/Daily Mail \citep{nallapati2016abstractive} and XSum \citep{narayan2018don}, confirms HeSum's high abstractness, compression ratio, and low redundancy, even when 
compared to these datasets.

\begin{table*}[th]
\centering
\scalebox{0.53}{
\begin{tabular}{llllrll}
\textbf{Phenomenon}         & \textbf{GPT-4} & \textbf{GPT-3.5} & \textbf{mLongT5} &  \multicolumn{1}{c}{\textbf{\begin{tabular}[c]{@{}c@{}}Example  error\\  in Hebrew\end{tabular}}}                                                                                                & \multicolumn{1}{c}{\textbf{\begin{tabular}[c]{@{}c@{}}Example error \\ translated into English\end{tabular}}}                                                                              & \multicolumn{1}{c}{\textbf{Explanation}}                                                                                                                                                                                      \\ \Xhline{6\arrayrulewidth}
\textbf{Repetition}         & \pz0              & \pz0                & \pz1            & \begin{tabular}[c]{@{}r@{}}

?\begin{cjhebrew}Myl' twyhl lwky 'wh M'h\end{cjhebrew}

\\
...?\begin{cjhebrew}Myl' twyhl lwky 'wh M'\end{cjhebrew}

\end{tabular} & \begin{tabular}[c]{@{}l@{}}Can he be violent?  If he \\  can be violent?\end{tabular}                & \begin{tabular}[c]{@{}l@{}}Duplication with subtle alterations.\end{tabular}                       \\ \cline{1-7}

\textbf{\begin{tabular}[c]{@{}l@{}} Copy  from\\ \textbf{Article}\end{tabular}    }         & \pz0              & \pz0                & \pz4            & \begin{tabular}[c]{@{}r@{}}

\begin{cjhebrew}\cjRL{h'M `ytwn'y ykwl lh.sy` }\end{cjhebrew}

\\
...\begin{cjhebrew}\cjRL{byqwrt pwmbyt `l}\end{cjhebrew}

\end{tabular} & \begin{tabular}[c]{@{}l@{}}Can a journalist make  \\  public criticism of...\end{tabular}                & \begin{tabular}[c]{@{}l@{}}The model-generated summary replicates a section of the original article.\end{tabular}                       \\ \cline{1-7}

\textbf{\begin{tabular}[c]{@{}l@{}} Non-alphabetic  \\ \textbf{omission}\end{tabular}    } 

& \pz1              & \pz0                & 14            & 

\begin{tabular}[c]{@{}r@{}}
\begin{cjhebrew}\cjRL{.s.h.s.hyM}\end{cjhebrew}\\

\end{tabular}
&  \begin{tabular}[c]{@{}l@{}}Chachachim \end{tabular}                                                                                                         & 

\begin{tabular}[c]{@{}l@{}}
 Missing diacritic -- it should be `\begin{cjhebrew}\cjRL{.hyM}\end{cjhebrew}\textquotesingle\begin{cjhebrew}\cjRL{.s|}\end{cjhebrew}\begin{cjhebrew}\cjRL{.h|}\end{cjhebrew}\textquotesingle\begin{cjhebrew}\cjRL{.s|}\end{cjhebrew}' instead of `\begin{cjhebrew}\cjRL{.s.h.s.hyM}\end{cjhebrew}'.\end{tabular}
                                                                              \\ \cline{1-7}

\textbf{\begin{tabular}[c]{@{}l@{}} Incorrect \\ \textbf{disambiguation}\end{tabular}    } 

& \pz1              & \pz1                & \pz2            & 

\begin{tabular}[c]{@{}r@{}}
\begin{cjhebrew}\cjRL{n'wmw /sl hdwd .twpz}\end{cjhebrew}...\\

\end{tabular}
&  \begin{tabular}[c]{@{}l@{}}Uncle Topaz's speech... \end{tabular}                                                                                                         & 

\begin{tabular}[c]{@{}l@{}}
 `\begin{cjhebrew}\cjRL{dwdw}\end{cjhebrew}'  is incorrectly interpreted as 
  `\begin{cjhebrew}\cjRL{dwd}\end{cjhebrew}' + `\begin{cjhebrew}\cjRL{/sl}\end{cjhebrew}' (`Uncle' + 'of'), \\ instead of  as  a man's name --  `\begin{cjhebrew}\cjRL{dwdw}\end{cjhebrew}', which is why the model \\ added a definite `\begin{cjhebrew}\cjRL{h}\end{cjhebrew}'
  to `\begin{cjhebrew}\cjRL{dwd}\end{cjhebrew}'. 
  \end{tabular}
                                                                              \\ \cline{1-7}   
\textbf{Hallucination}      & \pz3              & \pz3                & \pz2            & 
...\begin{cjhebrew}\cjRL{`yrb 't n.h}\end{cjhebrew}...

& ...involved Noah...                                                                                                                         & 

\begin{tabular}[c]{@{}l@{}} Noah is not a person mentioned in the article.      
\end{tabular}                                                                                                       \\ \cline{1-7}    
\textbf{Culture transfer}   & \pz1              & \pz1                & \pz0            & 

\begin{tabular}[c]{@{}r@{}}

,\begin{cjhebrew}tr.hbnh Nyypmqh tgyhnml\end{cjhebrew}...

\\
...\begin{cjhebrew}sdnrb \cjRL{nnsy}\end{cjhebrew}

\end{tabular}

& \begin{tabular}[c]{@{}l@{}}...to the campaign \\ leader-elect,  Nancy  \\ Brands...\end{tabular}                          & \begin{tabular}[c]{@{}l@{}}
The article refers to Nancy as a `he', but the summary uses feminine  \\ inflection (leader), probably due to  Nancy being a  common female \\ name in English.
\end{tabular}                                                                 \\ \cline{1-7}   
\textbf{Incorect gender}    & \pz6              & 11                & \pz1            & 

...\begin{cjhebrew}\cjRL{.hw/spwt b.hqyrtM}\end{cjhebrew}...

& \begin{tabular}[c]{@{}l@{}} ...reveal in their \\ investigation...\end{tabular} & \begin{tabular}[c]{@{}l@{}}

Gender inflection mismatch: `reveal'  (fem.) clashes with  `their'  \\(masc.).
\end{tabular}                                  
\\ \cline{1-7}                                                                                                            
\textbf{\begin{tabular}[c]{@{}l@{}} Incorrect definite \\ \textbf{(e.g., construct state)}\end{tabular}    } & \pz3              & \pz2               & \pz3            & 

...\begin{cjhebrew}\cjRL{hm/srd hm/sp.tyM pyrsM}\end{cjhebrew}...

&  \begin{tabular}[c]{@{}l@{}}The Ministry of the\\  Justice published...\end{tabular}                                                                                                     & \begin{tabular}[c]{@{}l@{}}

Definite articles on both words in `The Ministry of the Justice' \\ violate  Hebrew construct state rules.

\end{tabular}  
\end{tabular}
}
\caption{Error analysis comparing generated summaries from GPT-4, GPT-3.5, and mLongT5 based on 30 inputs.
}
\label{tab:Qualitive_analysis}
 \end{table*}

\section{Experiments}

\subsection{Experimental setup}

\paragraph{Models} To demonstrate the complexity of this task, we conducted an evaluation of two LLMs in a zero-shot setting: the GPT-4 model with 32K context window (version 0613), and GPT-3.5-turbo with 16K context (version  0613). 
To find the most effective prompt format, we tested on the HeSum validation set various prompting strategies, including translating parts of the prompt to English.
Ultimately, we adopted the English-translated approach \citep{brown2020language}, where both the instruction and input were translated. The output summaries are strictly in Hebrew.
Additionally, to address the limitations of available generative models for Hebrew, we fine-tuned the multilingual mLongT5 model \cite{uthus2023mlongt5} on the HeSum training set with two versions, base (2.37 GB) and large (4.56 GB). mLongT5 is a sequence-to-sequence model based on mT5 \cite{xue2020mt5} specifically designed for handling long sequences.
 Appendix \ref{appendix:mt5} includes the GPT models' prompting strategies experiments, and the mLongT5 training details.




\paragraph{Automatic Evaluation Metrics} To evaluate  the generated summaries with respect to the original texts, we used two standardly-used automatic metrics: ROUGE and BertScore. 

ROUGE \citep{lin2004rouge} is a widely-used metric in  summarization that measures n-gram overlap between generated summaries and human-written references. We calculated ROUGE-1 (unigrams), ROUGE-2 (bigrams), and ROUGE-L scores (longest common subsequence) to capture different levels of granularity.
However, n-gram  metrics such as ROUGE can struggle with capturing semantic similarity if paraphrases are used. To address this, we also employed BertScore \citep{zhang2019bertscore} with AlephBert \citep{seker2021alephbert} as its backbone. BertScore leverages the pre-trained language model to provide a more semantically meaningful evaluation of the summary.

\paragraph{Human Evaluation}
To validate the quality of  model-generated summaries for the HeSum task, seven  independent expert annotators evaluated a total of 186 summaries (62 per model) based on the same set of 62 reference articles. 
Annotators evaluated each summary using a 1-5 Likert scale \cite{likert1932technique} based on two key criteria: \emph{coherence}, which assessed the summaries' grammaticality and readability, and \emph{completeness}, which measured the degree to which they capture the main ideas of the articles.
To measure the consistency of the annotators' scores, we calculated  Krippendorff’s $\alpha$ \cite{krippendorff2018content} for an interval scale, and received an $\alpha$  score of 0.78 which indicates a good inter-annotator agreement rate.

\subsection{Results}
\label{sec:results}
 \paragraph{Quantitative Analysis} 
Table~\ref{tab:models_results} summarizes the quantitative evaluation results. While mLongT5 consistently achieved higher ROUGE scores, a metric focused on surface-level similarity, GPT-based models exhibited superior performance in BertScore, a semantic similarity metric, and in human evaluation scores that assess coherence and completeness. The consistently higher ROUGE scores of mLongT5 might be partially attributed to limitations in the ROUGE metric itself. ROUGE scores favor summaries that closely mimic the source text, even if they lack originality or fluency. Additionally, n-gram-based metrics like ROUGE may discount grammatically correct sentences that convey the required meaning even with morphological or lexical word variations, or changes in word order, compared to the source text.

Furthermore, when comparing ROUGE to human evaluation we found a negative correlation of human evaluation with ROUGE scores. We computed Pearson correlation coefficients (PCC) and found the coefficients to be around -0.16 with highly statistically significant p-values (less than $2.39 \times 10^{-5}$), indicating that higher ROUGE scores do not in actuality correspond to human evaluations of good summaries. Similarly, using Kendall's $\tau$ correlations resulted in negative values.  Further research is needed for developing automatic summarization metrics that correlate with human scores.

\paragraph{Qualitative Analysis}

Following the identification of key error categories, we conducted a comparative analysis by randomly selecting 30 summaries generated by each of the three models for the same set of 30 articles. For each model, we then quantified the occurrences of each identified phenomenon within the sampled summaries.
The results in Table~\ref{tab:Qualitive_analysis} reveal disparities between the errors of GPT-based models and those of the fine-tuned mLongT5 on various linguistic phenomena. 

The finetuned mLongT5 exhibits pronounced disruptions like repetition (3.33\%) and exact copy of sections from the articles (13.33\%), which weren't observed in the GPT-based results. However, the GPT-based models demonstrate errors in morphological phenomena specific to Hebrew, such as incorrect gender and wrong definiteness marking on {\em smixut}, indicating that the morphological richness of the language remains a challenge for these LLMs.  Additionally, known phenomena of GPT-based models such as “hallucinations”  \citep{cui2023holistic, guerreiro2023hallucinations} are also observed in our analysis, as is familiar from other languages.

\section{Conclusion}

This research seeks to fill a critical gap in the field of  LLMs assessment for generative creative tasks in MRLs, 
by presenting HeSum, a new dataset for Hebrew abstractive summarization, that includes 10K article-summary pairs sourced from professional journalists on Hebrew news websites. HeSum offers three key advantages:
high level of abstractness in summarization, distinct challenges specific to the Hebrew language, and a thorough empirical assessment of LLMs using this dataset.
By integrating these aspects, HeSum establishes itself as a valuable resource for researchers striving to push the boundaries of generative tasks, and specifically abstractive text summarization in Hebrew.

\section*{Limitations}

\paragraph{Evaluation} 
Metrics based on n-gram matching, such as BLEU \citep{papineni2002bleu}, ROUGE \cite{lin2004rouge}, and Meteor \cite{denkowski2014meteor}, are commonly used for evaluating summarization quality in English. However, these metrics can be problematic when applied to Hebrew text. Hebrew allows for more flexible word order compared to English. Additionally, its morphological richness entails that  the same concept can be expressed in multiple ways due to variations in prefixes, suffixes, and root conjugations.  
Furthermore, Hebrew has variations in spelling words due to missing vowels (\emph{Ktiv haser} vs. \emph{Ktiv male}). 
These factors can lead to n-gram-based metrics overlooking grammatically correct sentences in the generated summary that convey the full meaning even though they show slight differences.
Our findings in Section~\ref{sec:results}, which demonstrate a negative correlation between ROUGE scores and human evaluation scores, highlight the limitations of ROUGE evaluation in the context of Hebrew summarization.

\paragraph{Subset of LLMs} Although we aspired to evaluate HeSum on a broad range of large language models (LLMs), our current analysis is limited to only two generative models. This might overlook newer models offering potentially superior performance. 
Additionally, resource constraints prevented us from investigating the behavior of these models in few-shot settings. Having acknowledged that, the timeliness of this resource is uncompromized, as it can be used with contemporary and future models alike, to track advances on this challenge.

\paragraph{Open Access vs.\ Domain Focus}
HeSum predominantly comprises articles from news websites, which may bias models' success in this task towards news-style writing, and may not fully capture the linguistic diversity across different genres and domains. The reason for selecting these domains specifically stems from our ability to obtain a permissive license for the resource,  allowing open and free access by the community. However, the websites we have chosen -- “Shakuf”, “HaMakom”, and “The Seventh Eye” --  deviate from typical news platforms, offering a diverse range of topics that go beyond the typical content found on many popular news websites in Hebrew. This variety ensures that our dataset reflects a broader spectrum of real-world topics.

\paragraph{Dataset Scale vs.\ Quality}
In the realm of abstractive summarization, datasets like CNN/Daily Mail \citep{nallapati2016abstractive} and XSum \citep{narayan2018don} are commonly employed. These datasets utilize news articles from websites, treating the article content as the document and a corresponding field (often not explicitly intended as a summary) as the summary. However, this approach has faced criticism due to uncertainty about whether the chosen field truly represents a summary \cite{tejaswin-etal-2021-well}. An alternative approach involves human summarization, but this tends to result in smaller datasets (e.g., PriMock57, \citealp{papadopoulos2022}). To advance Hebrew NLP (and the study of generation in MRLs in general) beyond traditional classification tasks, there is a need for extensive generative datasets.  Given the current lack of viable alternatives within the NLP community, we have adopted a similar approach to XSUM, albeit with longer summaries. 
Additionally, collecting human-generated summaries in low-resource languages presents challenges, including the scarcity of crowdsourcing platforms that support Hebrew. To ensure quality, we meticulously reviewed 100 articles, their subheadings, and brief introductory sentences. Ultimately, we chose subheadings as our summary source because they provided more informative content, capturing additional details from the articles. Furthermore, we filtered out articles with subheadings containing very few tokens (10 or fewer) to ensure our summaries adequately represent the article content.

\section*{Ethics}
Following the generous permission of “Shakuf”, “HaMakom”, and “The Seventh Eye” --- organizations committed to independent journalism, media scrutiny, and transparency in Israel  --- we were granted the valuable opportunity not only to access and analyze their published articles but also to publish the data for broader research use. This unique collaboration fosters open access and empowers other researchers to build upon the data extracted from their articles and our findings within Hebrew abstraction summarization, expanding knowledge in this important field. Also, we are guaranteed not to have offensive language or hate speech in our data.  It should be borne in mind, however, that the opinions or biases reflected in these data may differ from other sources of information (news websites, social media, non-Hebrew news reports, and the like). So, the deployment of technology trained on this resource should be done with care.

\section*{Acknowledgements}
We wholeheartedly thank Naomi Niddam (Editor-in-Chief, “Shakuf”), Nir Ben-Zvi (CEO, “Shakuf” and “The Seventh Eye”), Shuki Tausig (Editor-in-Chief, “The Seventh Eye”), and Ohad Stone (Creative and digital projects, “Shakuf”) for granting us access to and permission to analyze and share their articles for broader research purposes.
We also thank Yoav Goldberg, Ido Dagan, Amir David Nissan Cohen, and the participants of the NLP seminar at Bar-Ilan University for their valuable feedback.
This research has been funded by a grant from
the European Research Council, ERC-StG grant
number 677352,  a grant by the Israeli Ministry of Science and Technology (MOST), and a KAMIN grant from the Israeli Innovation Authority, for which we are grateful. 
The first, second and last authors have been funded by the Israeli Science Foundation (grant ISF 670/23).

\bibliography{anthology}
\bibliographystyle{acl_natbib}

\appendix

\begin{table*}[t]
\centering
\begin{tabular}{lcccccccc}
\toprule
Dataset & \multicolumn{4}{c}{novel n-grams} &  CMP & RED (n=1) & RED (n=2) \\
\cmidrule{2-5}
       & n = 1 & n = 2 & n = 3 & n = 4 & & & & \\
\midrule
CNN/Daily Mail & 13.20 & 52.77 & 72.20  &  81.40&  90.90 &  13.73 & 1.10 \\
XSum   & 35.76 & 83.45 & 95.50 & 98.49 & 90.90 & \pz5.83  & 0.16 \\
HeSum  & 42.00    & 73.20   & 82.00 & 85.36 &  95.48 &  \pz4.83  & 0.10 \\
 \cdashline{1-9}
HeSum (morpheme-based)  & 17.41    & 48.02   & 67.51 & 76,86 &  95.57 &  25.90  & 2.72 \\
\bottomrule
\end{tabular}
\caption{HeSum's Intrinsic Evaluation compared to English Benchmarks (CNN/Daily Mail and XSum).}
\label{tab:Abstractness-morpheme}
\end{table*}

\section{The HeSum Dataset}

\paragraph{Collection Protocol}
Since the websites we collected (Shakuf, HaMakom, and The Seventh Eye) lack archives or RSS feeds, we developed a crawler to systematically navigate through pages, beginning from the homepage and exploring various article links. Leveraging their shared HTML structure, we could efficiently scrape the sites. We excluded pages without textual content, such as multimedia pages or those not in Hebrew. Additionally, articles with summaries of less than 10 tokens were filtered out, as they often lack sufficient detail to be a summary. In addition, all the articles were cleaned from Unicode characters or unrelated content.

\definecolor{lightgray}{RGB}{245,245,245} 
\noindent
\begin{minipage}[t][0.5\textheight]{0.5\textwidth} 
\begin{mdframed}[backgroundcolor=lightgray, linewidth=0pt]
\textbf{Coherence}
\begin{enumerate}
  \itemsep0pt 
  \item Very Incoherent: The summary is extremely confusing and lacks any clear connection between sentences.
  \item Incoherent: The summary is somewhat understandable.
  \item Somewhat Coherent
  \item Coherent
  \item Very Coherent
\end{enumerate}

\textbf{Completeness}
\begin{enumerate}
  \itemsep0pt 
  \item Very Incomplete: The summary lacks essential information and does not convey the main points effectively.
  \item Incomplete: The summary provides some information but misses key details.
  \item Somewhat Complete
  \item Complete
  \item Very Complete
\end{enumerate}
\end{mdframed}
\captionof{figure}{Evaluation Criteria} 
\label{figure:evaluation_Criteria}
\end{minipage}

\paragraph{Human Evaluation Details}

We collected annotations from seven volunteered participants aged 25 and above, all with at least one academic degree.
The participants were instructed to rate two parameters -- \emph{coherence} and \emph{completeness}, based on known criteria, as depicted in Figure \ref{figure:evaluation_Criteria}.
While completeness measures the extent to which the summary captures all the essential information from the source text, coherence is a more complex metric.  According to \citet{reinhart1980conditions}, coherence encompasses three core aspects: (i) cohesion, (ii) consistency, and (iii) relevance. While metrics like BertScore can also assess completeness, automatic evaluation of coherence remains a challenge \cite{maimon2023cohesentia}. Therefore, the measurement of coherence is evaluated in this work solely by humans.

\paragraph{Data Analysis}
Table~\ref{tab:Abstractness-morpheme} provides a quantitative analysis of HeSum's summarization characteristics, highlighting its challenges. The analysis utilizes two tokenization approaches: word-based (above the dashed line) and morpheme-based (below the dashed line). This distinction allows for a deeper examination of the dataset's abstractness and the influence of morphological features. As the table demonstrates, the number of unique vocabulary items (novel n-grams) decreases when using morpheme tokenization. However, HeSum still exhibits a higher uni-gram count compared to CNN/Daily Mail. This suggests that the task itself inherently involves a high degree of abstractness, and the morphological nature of the data presents an additional challenge.

\section{Models}
\label{appendix:mt5}

\begin{table}[t]
\centering
\resizebox{\columnwidth}{!}{%
\begin{tabular}{llllrrrr}
\toprule
\textbf{Model} & \textbf{Rouge1} & \textbf{Rouge2} & \textbf{RougeL} & \textbf{Epochs} & \textbf{Loss} \\ 
\midrule
\textbf{mLongT5-Base} & 18.62 & 8.68 & 15.92 & 18 & 2.15 \\
\textbf{mLongT5-Large} & 20.22 & 9.66 & 18.12 & 12 &  1.92 \\
\bottomrule
\end{tabular}%
}
\caption{mLongT5 performance on validation set and training details.}
\label{tab:mt5Long_details}
\end{table}

\begin{table*}[t]

\centering
\begin{tabular}{lcccccccc}
\toprule
Dataset & \multicolumn{4}{c}{novel n-grams} & CMP & RED (n=1) & RED (n=2) \\
\cmidrule{2-5}
       & n = 1 & n = 2 & n = 3 & n = 4 & & & & \\
\midrule
HeSum  & 42.00    & 73.20   & 82.00 & 85.36 &  95.48 &  \pz4.83  & \pz0.10 \\
 \cdashline{1-9}
GPT-4   & 47.24 & 80.35 & 91.37 & 95.92 &   91.89 & \pz8.14  & \pz0.68 \\
GPT-3.5 & 45.69 & 80.18 & 91.73  &  96.35 &  93.46 &  \pz7.53 & \pz0.83 \\
mLongT5-large  & \pz8.26 & 30.10 &  43.50   &  50.21 &  95.39 &  11.89  & \pz5.98 \\
mLongT5-base  & \pz7.21 & 28.77 &  42.06   &  49.46 &  92.25 &  15.74  & 10.25 \\

\bottomrule
\end{tabular}
\caption{Intrinsic Evaluation of Summarization. A Comparative Analysis of GPT-4, GPT-3.5, mT5 Models and the Hesum Dataset.}
\label{tab:model_abstractness}
\end{table*}

\paragraph{Fine-tunning mLongT5}
The HeSum corpus exhibits characteristics of long-form text, with an average document length of 2,747 tokens and a 90th percentile reaching 5,276 tokens. This extensive content poses challenges for the vanilla mT5 model, whose capacity for processing such lengths may be limited.
Consequently, we have fine-tuned the mLongT5 model \citep{uthus2023mlongt5}, which is suitable for handling long inputs. The paper presents results obtained with the base version of mLongT5, which is 2.37 GB. We are also releasing a larger model (4.56 GB).\footnote{\url{https://huggingface.co/biunlp/mT5LongHeSum-large}} The training regimen employed an 8-GPU A100 cluster for 36 hours for the large model, while the base model leveraged a single A100 GPU with 40 GB of memory. Early stopping, utilizing ROUGE-1 as the metric, was implemented to optimize the training process. Further details regarding model performance and implementation specifics are provided in Table \ref{tab:mt5Long_details}.

\begin{table}[t]
\resizebox{\columnwidth}{!}{
\begin{tabular}{llllrrrr}
\cline{1-7}
Model & prefix & input & output & \multicolumn{1}{l}{ROUGE-1} & \multicolumn{1}{l}{ROUGE-2} & \multicolumn{1}{l}{ROUGE-L}  \\ \cline{1-7}
GPT-3.5 & E     & E & E & 16.10  & 4.06 & 11.43                        \\
GPT-3.5 & H     & H & H & 16.34  & 4.26    & 11.69    \\
GPT-3.5 & E     & E & H & 12.80  & 2.30  &  11.00     \\                    
mLongT5     & NA & H & H & 17.47  & 7.56  & 14.68   \\                   
GPT-3.5 & E     & H & H & 17.08 & 4.95  & 12.46                \\
GPT-3.5 & H     & H & E & 14.35 & 3.13  & \pz9.89                      \\
GPT-3.5 & E     & H & E & 14.31  & 3.11  & 10.40                   \\ 
GPT-3.5 & H     & E & H & 15.90  & 4.23  & 10.80                       \\

\end{tabular}%
}

\caption{Testing different configurations of language prompting to find the best configuration to evaluate GPT-3.5. 'H' denotes Hebrew and 'E' denotes English. 'prefix' is the instruction to the model, 'input' is the article itself, and the output is the desired summarization language. }
\label{tab:GPT_language_config}
\end{table}

\paragraph{Prompting GPT-based models}
Here, we leverage the translate-English approach, suggested by \citep{shi2022language} and \citep{ahuja2023mega}, which translates instances from target languages into English before prompting. 
We decompose the prompt task into three parts: (i) the input article (ii) the instruction (prefix), and (iii) the output. All three parts could be done in either Hebrew or English for the HeSum task. In our experiment, Google Translate API (2023, \citealp{googletranslateapi2023}) handled the translation of prompts (input and/or prefix) from Hebrew to English and the translated outputs back to Hebrew for analysis.
Testing GPT-3.5 on different configurations of language prompting in the HeSum validation set, we found that the best prompt-language configuration is English-English-English (Table \ref{tab:GPT_language_config}). We then applied this prompting strategy to both GPT-3.5 and GPT-4 on the test set. The prompt we used depicted in Figure \ref{figure:english_prompt}.


\definecolor{lightgray}{RGB}{245,245,245} 
\noindent
\begin{minipage}[0.7\textheight]{0.5\textwidth} 
\end{minipage}
\hfill 
\begin{mdframed}[backgroundcolor=lightgray, linewidth=0pt]
You are a genius summarizer. Your task is to summarize the main points of the following text. Please follow these instructions step by step:
\begin{enumerate}
  \item The summary should be concise, consisting of up to 3 sentences.
  \item If there are several main topics, create a separate sentence for each topic.
  \item The output should be in English.
\end{enumerate}
\end{mdframed}
\captionof{figure}{The prompt we used for the GPT-based models} 
\label{figure:english_prompt}

\section{Implementation Details}

For the intrinsic evaluation of the dataset, we created a Jupyter notebook which computes the different metrics. For computing the n-grams, we used the NLTK package,\footnote{\url{https://pypi.org/project/nltk/}} and for loading and processing the data, we used NumPy\footnote{\url{https://pypi.org/project/numpy/}} and Pandas.\footnote{\url{https://pypi.org/project/pandas/}} For evaluation of the different models, we used the most common ROUGE package for non-English papers,\footnote{\url{https://github.com/csebuetnlp/xl-sum/tree/master/multilingual_rouge_scoring}}
 and the HuggingFace implementation of Transformers for BertScore.\footnote{\url{https://pypi.org/project/transformers/}}

\section{Additional Models Performance Analysis}
Table \ref{tab:model_abstractness} presents the intrinsic evaluation results for the models, corresponding to the metrics introduced in Section \ref{sec:data_analysis}. Notably, GPT-based models generate text with greater abstractness, as evidenced by their higher count of novel n-grams compared to the fine-tuned mT5. This finding aligns with mT5's tendency towards repetitive generation, which is further supported by its high RED score and by the qualitative analysis presented in Table  \ref{tab:Qualitive_analysis}.

\end{document}